\pdfoutput=1
\documentclass[11pt]{article}
\usepackage{emnlp2021}
\usepackage{times}
\usepackage{latexsym}
\usepackage[T1]{fontenc}
\usepackage[utf8]{inputenc}
\usepackage{microtype}

\usepackage{braket}
\usepackage{subfigure}
\usepackage[export]{adjustbox}
\usepackage{microtype}
\AtBeginDocument{
  \let\mathbb\relax
  \DeclareMathAlphabet{\mathbb}{U}{msb}{m}{n}
}
\title{Neural News Recommendation with Collaborative News Encoding and Structural User Encoding}

\author{
  Zhiming Mao$^{1,2}$, Xingshan Zeng$^{3}$, Kam-Fai Wong$^{1,2}$ \\
  $^1$The Chinese University of Hong Kong, Hong Kong, China \\
  $^2$MoE Key Laboratory of High Confidence Software Technologies, China \\
  $^3$Huawei Noah’s Ark Lab, China \\
  $^{1,2}$\texttt{\{zmmao,kfwong\}@se.cuhk.edu.hk} \quad $^3$\texttt{zeng.xingshan@huawei.com} \\
}
\date{}

\begin{document}
\maketitle
\begin{abstract}
Automatic news recommendation has gained much attention from the academic community and industry. Recent studies reveal that the key to this task lies within the effective representation learning of both news and users. Existing works typically encode news title and content separately while neglecting their semantic interaction, which is inadequate for news text comprehension. Besides, previous models encode user browsing history without leveraging the structural correlation of user browsed news to reflect user interests explicitly. In this work, we propose a news recommendation framework consisting of collaborative news encoding~(CNE) and structural user encoding~(SUE) to enhance news and user representation learning. CNE equipped with bidirectional LSTMs encodes news title and content collaboratively with cross-selection and cross-attention modules to learn semantic-interactive news representations. SUE utilizes graph convolutional networks to extract cluster-structural features of user history, followed by intra-cluster and inter-cluster attention modules to learn hierarchical user interest representations. Experiment results on the MIND dataset validate the effectiveness of our model to improve the performance of news recommendation\footnote{Our code is released at \href{https://github.com/Veason-silverbullet/NNR}{https://github.com/Veason-silverbullet/NNR}}.
\end{abstract}

\section{Introduction}\label{introduction}
Online news applications, such as CNN News and MSN News, have become more and more people's first choices to obtain the latest news~\citep{google_news_recommendation}. With a deluge of news generated every day, an efficient news recommendation system should push relevant news to users to satisfy their diverse personalized interests~\citep{information_overload}.

From the perspective of representation learning ~\citep{representation_learning}, existing works mainly study how to effectively encode news and users into discriminative representations~\citep{DAE_RNN, NAML, FIM}. News encoders typically extract semantic representations of news from the textual spans (e.g., news title and content). User encoders are employed to learn the representation of a user from her browsing history. News recommendation models predict the matching probabilities between candidate news and users by measuring the similarity of their representations.

\begin{figure}[t]
\centering
\includegraphics[width=77mm]{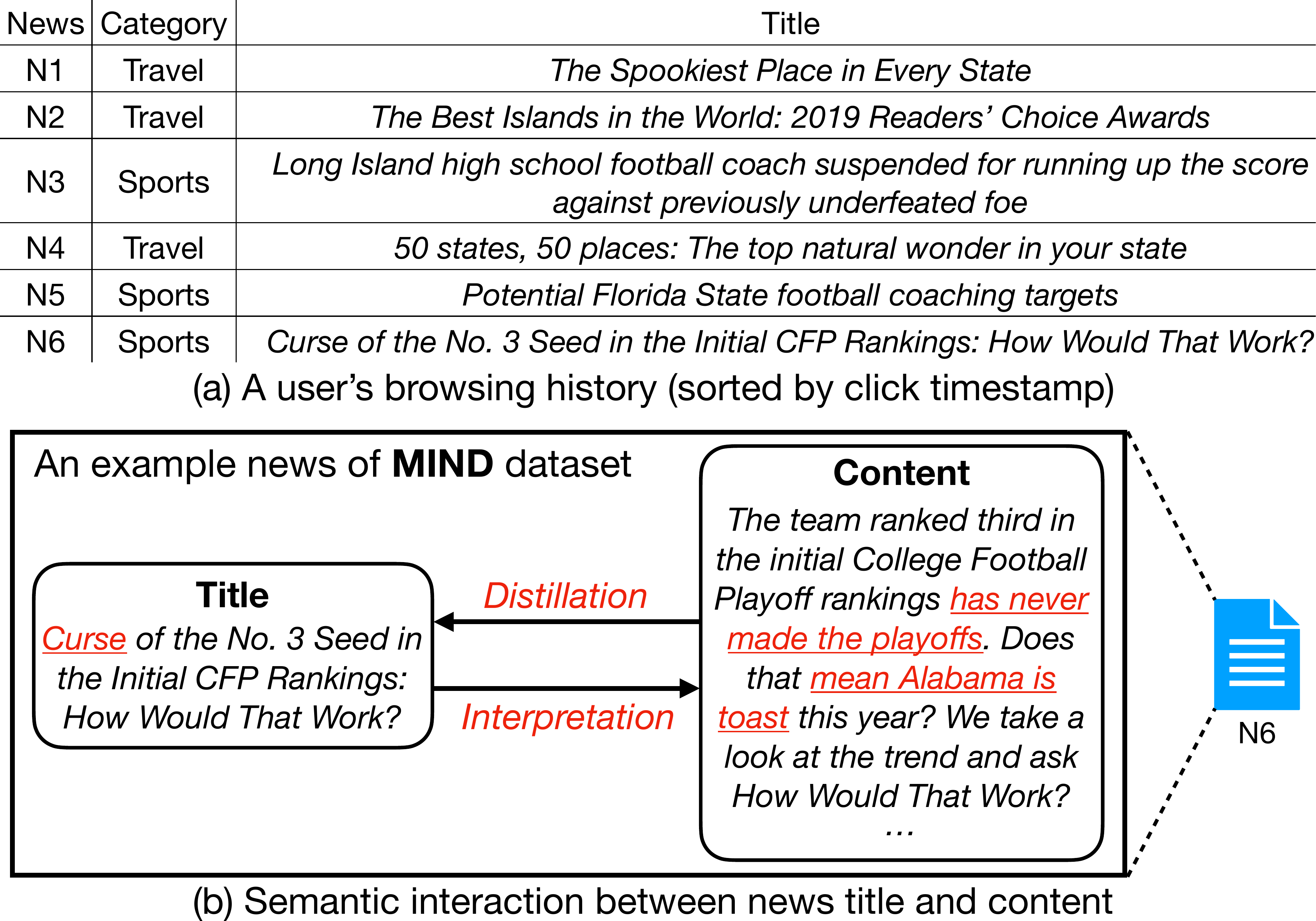}
\captionsetup{font=10pt}
\caption{
(a) An example of user browsing history. (b) An example of news title-content semantic interaction.
}
\label{fig:news_user}
\end{figure}
Existing news recommendation models typically encode news title and content separately and encode users' browsing histories without explicit structural modeling. We argue that these encodings restrict the power of the news and user representations. To enhance news and user encoding, this work is established based on the two aspects of news and user representation learning:

($1$) \textit{Encoding the semantic interaction between news title and content}: Title and content play different roles in news, but they are complementary. News title distills the information of content, while content interprets the details of title, as shown in Figure~\ref{fig:news_user}(b). Previous works treat news title and content as two separate textual features, leading to a ``\textit{semantic encoding dilemma}''. This dilemma is dyadic as: (\romannumeral 1) Although a news title is much shorter than content, the performance of title-encoding is empirically better than content-encoding~\citep{MIND}. This can be attributed to the crucial information that the human-summarized title naturally represents; (\romannumeral 2) News titles are always subjective and rhetorical to attract potential readers. This leads to a severe textual \textit{data sparsity} problem. News titles with \textit{unseen terminology}, \textit{metaphor} and \textit{ambiguity} make it difficult to comprehend news with limited title wording~\citep{NLU}. For example in Figure~\ref{fig:news_user}(b), the word ``\textit{curse}'' is a \textit{metaphor}, which cannot be resolved by the training corpus or the title itself, due to its unique semantic occurrence. News encoders must turn to the content to interpret the semantics of the word ``\textit{curse}''~(i.e., ``\textit{has never made the playoffs}'' and ``\textit{mean Alabama is toast}''). However, news encoders proposed by previous works either extract features solely from the title, or encode title and content separately, then perform concatenation or attention fusion on them~\citep{DAN, NAML}. Such separate encodings of title and content without leveraging their semantic interaction are inadequate for news text comprehension.

($2$) \textit{Encoding the user-interest-news correlation with hierarchical cluster-structure}: While a user usually has diverse interests in news topics, her browsed news with the same topic is often linked by some logical correlation. For example in Figure~\ref{fig:news_user}(a), the news \textit{N3}, \textit{N5} and \textit{N6} are labeled as sports news and logically correlated with the topic ``\textit{football}'', forming a virtual user interest cluster. None of the single news can precisely represent the overall user interest in ``\textit{football}''. However, refined user interest in ``\textit{football}'' can be encoded by aggregating the news \textit{N3}, \textit{N5} and \textit{N6}. With aspects of user interests encoded within specific clusters, overall user representations can be aggregated by leveraging the correlation among interest clusters. Previous works typically formulate user history as an ordered linear sequence of news. Based on this sequential formulation, recurrent neural networks \citep{DAE_RNN, LSTUR} and attention networks \citep{DAN, NAML, NPA, NRMS} are proposed to encode user history. These encoding methods viewing user history as a sequence of news cannot explicitly model the hierarchical \textit{user-interest-news} correlation. Compared to linear sequences, hierarchical clusters are more suitable to represent a user's diverse interests. User history can be structurally formulated into certain interest clusters, as correlated news shares information in a specific interest cluster. Encoding user history with hierarchical cluster-structure is more precise to represent the correlation of news and user interests.

To address the above issues, in this work, we propose collaborative news encoding~(CNE) and structural user encoding~(SUE) to learn semantic-interactive news representations and hierarchical user representations. We conduct experiments on the MIND dataset~\citep{MIND}, showing the encoding effectiveness of our proposed model. Experiments and further analyses validate that (\romannumeral 1) CNE can enhance news encoding by exploiting the word-level semantic interaction between news title and content with cross-selective and cross-attentive mechanisms; (\romannumeral 2) SUE utilizes hierarchical cluster graphs to model the correlation of a user's browsed news, which can extract more precise user interest representations; (\romannumeral 3) our model significantly outperforms existing state-of-the-art news recommendation models on the real-world MIND dataset.

\section{Related Work}
News recommendation is not only an important research task in NLP~\citep{MIND} but also a core component of industrial personalized news service~\citep{DAE_RNN}. Conventional collaborative filtering (CF) approaches~\citep{CF} exploit the interaction relationship between news and users. Since news only lasts for a short period, CF-based methods suffer from severe \textit{cold-start} problem. To tackle this, content-based methods used handcrafted features to encode news and users \citep{handcraft2,handcraft3,handcraft4}. In recent years, deep neural models have achieved superior performance in news recommendation. Many studies pinpointed that this improvement came from the fine-grained news and user representations, which were extracted by deep neural networks~\citep{NAML,NRMS,FIM}.

For news representation learning, existing works used convolutional neural networks (CNN) \citep{LSTUR}, knowledge-aware CNN \citep{DKN}, personalized attention networks \citep{NPA}, and multi-head self-attention networks \citep{NRMS} to extract features from news title text as news representations. \citet{DAN} employed parallel CNNs to encode news title and content respectively and then concatenated them into a unified representation. \citet{NAML} encoded news title and content separately and incorporated them with multi-view attention.

For user representation learning, \citet{DAE_RNN} used GRU to encode user history by ordered timestamp. \citet{LSTUR} utilized RNN to learn short-term user representations from the browsing history, combined with long-term user embeddings. Various attention networks are also widely used to attend to important news in user history~\citep{NAML, NPA, DAN}. \citet{NRMS} employed multi-head self-attention~\citep{transformer} to capture deep interaction of user browsed news. These works formulated user history as an ordered linear sequence of news, to which recurrent or attention models were applied without modeling the structural correlation of user browsed news. \citet{GNUD} formulated news and user jointly with a bipartite graph and disentangled user preferences with routing mechanism, which however implicitly relied on the manually-set latent preference factor.

\section{Methodology}
\begin{figure*}[t]
\centering
\includegraphics[height=73.98mm]{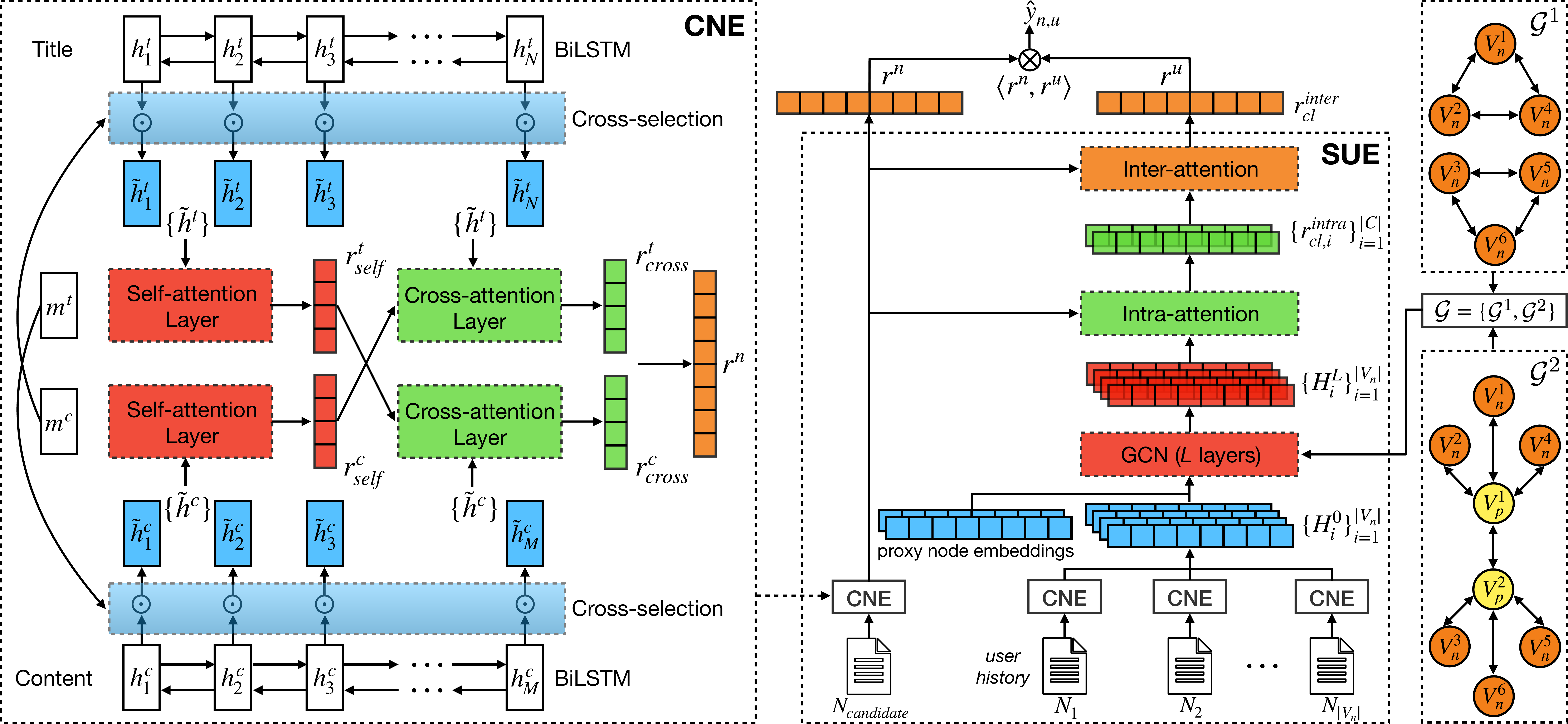}
\captionsetup{font=10pt}
\caption{
The overall architecture of our model. The graph construction is based on the user history in Figure~\ref{fig:news_user}(a).
}
\label{fig:model}
\end{figure*}

Our model is composed of the \textbf{C}ollaborative \textbf{N}ews \textbf{E}ncoding (CNE) module presented in Section~\ref{ssec:CNE} and \textbf{S}tructural \textbf{U}ser \textbf{E}ncoding (SUE) module presented in Section~\ref{ssec:SUE}. CNE and SUE extract representations of candidate news and users respectively. The overall model architecture is illustrated in Figure~\ref{fig:model}. Finally, Section~\ref{ssec:CPMT} will describe the click predictor and details of model training.

\subsection{Collaborative News Encoding}\label{ssec:CNE}
\subsubsection{Cross-selective Encoding}
The news encoder is employed to learn semantic news representations from news title and content. Given the title word sequence $x^{t}=[x^t_1,x^t_2,...,x^t_N]$ and content word sequence $x^{c}=[x^c_1,x^c_2,...,x^c_M]$, they are mapped to word embeddings $W^{t}=[w^t_1,w^t_2,...,w^t_{N}]$ and $W^{c}=[w^c_1,w^c_2,...,w^c_{M}]$, where $N$ and $M$ are the word sequence lengths. For simplicity, we only formulate the title encoding part of our model~(denoted by superscript $t$). The content encoding formula is symmetric as a counterpart~(denoted by superscript $c$) and omitted.

First, two parallel bidirectional LSTMs are employed to extract the sequential features from the title and content word embeddings respectively.
\begin{equation}\label{eq1}
    \overrightarrow{h}^{t}_{i}=\overrightarrow{LSTM}^{t}(w^{t}_{i}, \overrightarrow{h}^{t}_{i-1}, \overrightarrow{c}^{t}_{i-1})
\end{equation}
\begin{equation}\label{eq2}
    \overleftarrow{h}^{t}_{i}=\overleftarrow{LSTM}^{t}(w^{t}_{i}, \overleftarrow{h}^{t}_{i+1}, \overleftarrow{c}^{t}_{i+1})
\end{equation}
where $\{h^t\}$ and $\{c^t\}$ are LSTM hidden states and cell states. The \textit{i}-th title sequential feature is fused as $h^t_i=[\overrightarrow{h}^{t}_{i};\overleftarrow{h}^{t}_{i}]$, where $[\cdot;\cdot]$ denotes vector concatenation. We consider the global semantic information of title~(content) preserved in its LSTM cell states and concatenate the last forward $\overrightarrow{c}^{t}_{N}$ and backward $\overleftarrow{c}^{t}_{1}$ as the semantic memory vector $m^{t}$.
\begin{equation}\label{eq3}
m^t=[\overrightarrow{c}^{t}_{N};\overleftarrow{c}^{t}_{1}]
\end{equation}

To facilitate semantic interaction between title and content, we design a gated cross-selective network, inspired by \citet{selective-mechanism}. Concretely, we utilize the semantic memory vector $m^{c(t)}$ to perform feature recalibration~\citep{SEnet} on the sequential features $\{h^{t(c)}\}$ by a sigmoid gate function. The motivation behind this gate function is to utilize the memory vector of content~(title) $m^{c(t)}$ to cross-select important semantic information from the $i$-th title~(content) sequential feature $h^{t(c)}_i$.
\begin{equation}\label{eq4}
    Gate^{t}_{i}=\sigma(W_{g}^{h}h^{t}_i+W_{g}^{m}m^{c}+b_g)
\end{equation}
\begin{equation}\label{eq5}
    \tilde{h}^{t}_i=Gate^{t}_{i}\odot{h^{t}_i}
\end{equation}
where $\sigma$ is sigmoid activation, $\odot$ denotes element-wise multiplication. $\tilde{h}^{t}_i$ is the cross-selective feature of the \textit{i}-th title sequential feature $h^t_i$ interacting with the content memory vector $m^c$. It is the first stage of collaborative title-content semantic interaction.

\subsubsection{Cross-attentive Encoding}
Based on the cross-selective sequential feature $\{\tilde{h}\}$, a two-phase attention module is designed to learn cross-attentive representations of title and content. First, we employ self-attention layers to learn the self-attentive representation of the sequential $\{\tilde{h}\}$.
$$\label{_eq6}
    \alpha^{t}_{self}=softmax(v^\mathsf{T}tanh(W\tilde{h}^{t}+b))
$$
\begin{equation}\label{eq6}
    r^{t}_{self}=\sum_{i=1}^{N}{\alpha^{t}_{self,i}\tilde{h}^{t}_i}
\end{equation}

Then we employ the self-attentive representation $r_{self}$ as a query, and the $\{\tilde{h}\}$ as key-value pairs to build cross-attention layers\footnote{Practically, we employ the scaled dot-product attention proposed by~\citet{transformer}. $Attention(Q,K)=softmax(\frac{\bar{Q}\bar{K}^{T}}{\sqrt{d}})$, where $\bar{Q}=QW^Q$, $\bar{K}=KW^K$. The same attention functions are also applied in Eq.~(\ref{eq10}) and (\ref{eq12}).}. It is the second stage of collaborative title-content semantic interaction.
$$\label{_eq7}
    \alpha^{t}_{cross}=Attention(r^{c}_{self}, \{\tilde{h}^{t}\})
$$
\begin{equation}\label{eq7}
    r^{t}_{cross}=\sum_{i=1}^{N}{\alpha^{t}_{cross,i}\tilde{h}^{t}_i}
\end{equation}

We compute element-wise summation~(denoted as $\oplus$) of the self-attentive representation $r_{self}$ and cross-attentive representation $r_{cross}$ to derive the title and content semantic-interactive representations, i.e., $r^t$ and $r^c$. Finally, we concatenate $r^t$ and $r^c$ as the collaborative news representation $r^n$.
\begin{equation}\label{eq8}
    r^n=[r^{t};r^{c}]=[r^{t}_{self}\oplus{r^{t}_{cross}};r^{c}_{self}\oplus{r^{c}_{cross}}]
\end{equation}

\subsection{Structural User Encoding}\label{ssec:SUE}
\subsubsection{Cluster-based Encoding of User History}\label{sssec:GCN}
The user encoder is employed to learn user interest representations from their browsing histories. To formulate the cluster-structure of user interests, we construct a hierarchical cluster graph in two steps:

\textbf{Intra-cluster Subgraph $\mathcal{G}^1$}. We construct an original cluster graph\footnote{\href{https://en.wikipedia.org/wiki/Cluster\_graph}{https://en.wikipedia.org/wiki/Cluster\_graph}} with the topic category label of news~(e.g., ``\textit{Sports}'' and ``\textit{Travel}'' in Figure~\ref{fig:news_user}(a)). We build the subgraph $\mathcal{G}^1=(V_n, E_n)$ by treating the browsed news as nodes $\{V_n\}$ and adding bidirectional edges $\{E_n\}$ to those nodes, which share the same category labels. Each news node of $\{V_n\}$ is associated with its embedding $r^n$ in Eq.~(\ref{eq8}). Each cluster contains multiple browsed news with a specific topic, reflecting an aspect of user interests.

\textbf{Inter-cluster Subgraph $\mathcal{G}^2$}. Besides intra-cluster refinement of user interests, modeling inter-cluster correlation is also essential to leverage the overall information of user history. For each cluster $C_i$ in $\mathcal{G}^1$, we add a new cluster proxy node $V_p^i$. We build the subgraph $\mathcal{G}^2=(\{V_n,V_p\}, \{E_p^1,E_p^2\})$ by adding bidirectional edges $\{E_p^1\}$ to those news nodes $\{V_n\}$ and proxy nodes $\{V_p\}$ within the same clusters and fully connecting $\{V_p\}$ by bidirectional edges $\{E_p^2\}$. The node embedding of $\{V_p\}$ is initiated as zero-embedding $r^p$. News node information among clusters aggregates via cluster proxy nodes.

The hierarchical cluster graph $\mathcal{G}$ consists of intra- and inter-cluster subgraphs: $\mathcal{G}=\{\mathcal{G}^1,\mathcal{G}^2\}$. With $d$-dimensional node embedding vectors $\{r^n_i\}_{i=1}^{|V_n|}$ and $\{r^p_i\}_{i=1}^{|V_p|}$, we define the history feature matrix as $H^{0}=[r^n;r^p]\in{\mathbb{R}^{(|V_n|+|V_p|)\times{d}}}$. For graph $\mathcal{G}$, we denote its normalized adjacency matrix as $\tilde{A}$ and degree matrix as $\tilde{D}$. We use graph convolutional networks~(GCN)~\citep{GCN} to extract structural representations of user history. To mitigate the over-smoothing issue of deep GCN~\citep{over-smoothing}, we add residual connections to adjacent GCN layers, following \citet{ResGCN}.
\begin{equation}\label{eq9}
    H^{l+1}=ReLU(\tilde{D}^{-\frac{1}{2}}\tilde{A}\tilde{D}^{-\frac{1}{2}}H^{l}W^{l})+H^{l}
\end{equation}
where $W^l$ is a trainable matrix. The GCN extracts structural features on graph $\mathcal{G}$, refining specific user interest representations within clusters and aggregating overall user history information among clusters. We train GCN of $L$ layers and derive the structural user history representation from the news node embeddings as $r^{h}=\{H_i^L\}_{i=1}^{|V_n|}\in{\mathbb{R}^{|V_n|\times{d}}}$.

\subsubsection{Intra-cluster Attention}\label{sssec:intra-att}
Given the $|C|$ interest clusters implied by $|V_n|$ user's browsed news, there are $|C_i|$ news in cluster $C_i$. The structural user history representation $r^h$ can be viewed as $r^h=\{r^h_{i}\}^{|C|}_{i=1}=\Big\{\{r^h_{i,j}\}^{|C_i|}_{j=1}\Big\}^{|C|}_{i=1}$. To derive intra-cluster features associated with candidate news, we design an intra-cluster attention layer, regarding the candidate news representation $r^{n}_{can}$ as a query, and the $j$-th intra-cluster feature $r^{h}_{i,j}$ of cluster $C_i$ as a key-value pair.
\vskip -0.625em
$$\label{_eq10}
    \alpha^{intra}_{i}=Attention(r^{n}_{can}, \{r^{h}_{i}\})
$$
\vskip -1.2em
\begin{equation}\label{eq10}
    r^{intra}_{cl,i}=\sum_{j=1}^{|C_i|}{\alpha^{intra}_{i,j}r^{h}_{i,j}}
\end{equation}

The intra-cluster feature $r^{intra}_{cl,i}$ attends to the node-level features $\{r^h_{i,j}\}_{j=1}^{|C_{i}|}$ of cluster $C_i$, associated with the candidate news representation $r^n_{can}$. The $r^{intra}_{cl,i}$ refines the $i$-th user interest representation within the cluster $C_i$ of graph $\mathcal{G}$.

\subsubsection{Inter-cluster Attention}\label{sssec:inter-att}
Before inter-cluster modeling, a nonlinear transformation is performed to project the $r^{intra}_{cl,i}$, which is originally a linear combination of node-level features in cluster $C_i$, into cluster-level feature spaces.
\vskip -6mm
\begin{equation}\label{eq11}
    \tilde{r}^{intra}_{cl,i}=ReLU(\tilde{W}r^{intra}_{cl,i}+\tilde{b})+r^{intra}_{cl,i}
\end{equation}
\vskip -0.25em

To derive inter-cluster features associated with candidate news, we design an inter-cluster attention layer, regarding the candidate news representation $r^{n}_{can}$ as a query, and the $i$-th intra-cluster feature $\tilde{r}^{intra}_{cl,i}$ of graph $\mathcal{G}$ as a key-value pair.
\vskip -1.4em
$$\label{_eq12}
    \alpha^{inter}=Attention(r^{n}_{can}, \{\tilde{r}^{intra}_{cl}\})
$$
\vskip -1.6875em
\begin{equation}\label{eq12}
    r^{inter}_{cl}=\sum_{i=1}^{|C|}{\alpha^{inter}_{i}\tilde{r}^{intra}_{cl,i}}
\end{equation}

The inter-cluster feature $r^{inter}_{cl}$ attends to the cluster-level features $\{\tilde{r}^{intra}_{cl,i}\}_{i=1}^{|C|}$ of graph $\mathcal{G}$, associated with the candidate news representation $r^n_{can}$. With intra-cluster and inter-cluster attention, $r^{inter}_{cl}$ hierarchically aggregates user interest representations within the cluster graph $\mathcal{G}$. $r^{inter}_{cl}$ is adopted as the user representation $r^u$, i.e., $r^u=r^{inter}_{cl}$.

\subsection{Click Predictor and Model Training}\label{ssec:CPMT}
Given the news and user representations $r^n$ and $r^u$, the click predictor is employed to predict the probability that user \textit{$u$} clicks on news \textit{$n$}. Motivated by the previous works~\citep{DKN,NAML}, we compute the dot-product $\hat{y}_{n,u}$ of $r^n$ and $r^u$, i.e., $\hat{y}_{n,u}=\left\langle r^n,r^u \right\rangle$, as the unnormalized matching score of news \textit{$n$} and user \textit{$u$}.

Following common practice of previous works \citep{DSSM, FIM}, we employ negative sampling strategy to model training. For each user click-impression, i.e., the $i$-th impression log that user \textit{$u$} had clicked on news \textit{$n$}, we compute its unnormalized matching score as $\hat{y}^{+}_i$. Besides, we randomly sample $K$ pieces of news, which are not clicked by the user \textit{$u$}. Unnormalized matching scores $\hat{y}^-_{i,j}$ are computed for these $K$ negative samples, where $j=1,...,K$. By such means, it can be reformulated as a $(K+1)$-way classification problem. We employ softmax function to derive the normalized matching probabilities and sum up the negative log-likelihood of positive samples over the training dataset $\mathcal{D}$, as model training loss $\mathcal{L}$.
\vskip -6.25mm
\begin{equation}\label{eq13}
    \mathcal{L}=-\sum_{i=1}^{|\mathcal{D}|}{log\frac{exp(\hat{y}^+_i)}{exp(\hat{y}^+_i)+\sum_{j=1}^{K}{exp(\hat{y}^-_{i,j})}}}
\end{equation}
\vskip -3.125mm

\begin{table}[t]
\begin{center}
\setlength{\tabcolsep}{1.6mm}
\fontsize{8.25}{11}\selectfont
\begin{tabular}{|c|c|c|c|}
\hline 
\# users&200,000 & \# users in train set & 189,532\\
\hline
\# news&78,520 & \# news in train set & 75,858\\
\hline
\# training logs & 594,433 & \# positive samples & 902,330\\
\hline
Avg. title len & 11.67 & Avg. content len & 41.01\\
\hline
\end{tabular} 
\end{center}
\captionsetup{font=10pt}
\caption{Statistics of the $200$K-MIND dataset.}
\label{tables:statistics}
\end{table}
\section{Experiment Setup}
\subsection{Dataset and Experiment Settings}
We conduct experiments on the MIND dataset~\citep{MIND}. MIND is a large-scale English news recommendation dataset built from real-world MSN news and anonymized user click logs\footnote{\href{https://msnews.github.io}{https://msnews.github.io}}. Since the MIND is quite large-scale, following existing works~\citep{NAML, NPA, NRMS, FIM}\footnote{These works used the MSN news dataset with 10K sampled users, as training on the full MIND dataset with 1 million users is very expensive in GPU time.}, we randomly sample $200$K users' click logs out of $1$ million users from the user behavior logs of MIND training and validation sets. Since the MIND test set is not labeled, we half-split the original validation set into experimental validation and test sets. We employ the abstract column texts in MIND as the news content texts\footnote{Detailed MIND dataset format at \href{https://github.com/msnews/msnews.github.io/blob/master/assets/doc/introduction.md}{https://github.com/msn \\ ews/msnews.github.io/blob/master/assets/doc/introduction.md}}. Detailed statistics of the $200$K-MIND dataset are shown in Table~\ref{tables:statistics}.

We truncate news title and content with the maximum length of $32$ and $128$ respectively. The number of news in user browsing history is capped at $50$. Following \citep{LSTUR, NAML, NPA, NRMS}, we perform negative sampling with the sampling ratio $K$ of $4$ (see Section~\ref{ssec:CPMT}). The word embedding is initialized from the pretrained $300$-dimensional Glove embedding~\citep{glove}. The number of GCN layers in SUE is set as $L=4$~(investigated in Section~\ref{GCN_layer_num}). We use Adam optimizer~\citep{Adam} with the learning rate of $1$e-$4$ to train our model with the dropout rate of $0.2$. The area under the ROC curve (AUC), mean reciprocal rank (MRR), and normalized discounted cumulative gain (nDCG@$5$ and nDCG@$10$) are adopted as ranking metrics to evaluate model performance. We set the batch size to $64$ and conduct early stopping if the validation AUC score had not improved over $5$ epochs. We independently repeat each experiment for $10$ times and report the average performance scores.

\subsection{Comparison Methods}
We compare our model with state-of-the-art general and news-specific recommendation methods.

\textbf{General Recommendation Methods.}
General methods utilize handcrafted features to learn news and user representations. We use the TF-IDF features extracted from news and user history with the one-hot news and user IDs as input features for the experiments. The general methods include ($1$) \textbf{LibFM}~\citep{LibFM}, a factorization machine estimating the sparse feature interaction between news and users; ($2$) \textbf{DSSM}~\citep{DSSM}, a deep structured semantic model, regarding the user history as a query and candidate news as key documents; ($3$) \textbf{Wide\&Deep}~\citep{WD}, a framework consisting of wide channels with a linear model and deep channels with a neural model.

\textbf{Neural News Recommendation Methods.}
We compete with the state-of-the-art neural models, which are specifically designed for news recommendation, including ($1$) \textbf{DAE-GRU}~\cite{DAE_RNN}, encoding news with a denoising autoencoder and users with a gated recurrent network; ($2$) \textbf{DFM}~\citep{DFM}, a deep fusion model using multi-channel inception blocks to capture various interaction among news features; ($3$) \textbf{DKN}~\citep{DKN}, utilizing knowledge-aware CNNs to fuse knowledge encoding and textual encoding of news title; ($4$) \textbf{LSTUR}~\citep{LSTUR}, encoding news title with a CNN network, while jointly modeling long-term user preferences and short-term user interests with a GRU network; ($5$) \textbf{NAML}~\citep{NAML}, utilizing CNN networks to encode title and content texts, while encoding the category and subcategory topics with dense layers. The text and topic representations are incorporated by multi-view attention. NAML uses an attention network as its user encoder; ($6$) \textbf{NPA}~\citep{NPA}, attending to important words and news articles by personalized attention networks built with user embeddings; ($7$) \textbf{NRMS}~\citep{NRMS}, utilizing effective multi-head self-attention networks~\citep{transformer} to extract fine-grained representations from the news title and user history respectively; ($8$) \textbf{FIM}~\citep{FIM}, encoding news titles with dilated convolution networks and modeling the interaction between candidate news and user history with $3$D convolutional matching networks.

\textbf{Variants of Our Model.} To further verify the efficacy of our model design, we also experiment with the ablation variants of our model by respectively removing the cross-selection module (\textbf{CNE w/o CS}), cross-attention module (\textbf{CNE w/o CA}), GCN layers (\textbf{SUE w/o GCN}), and hierarchical cluster attention module~(\textbf{SUE w/o HCA}).

\section{Experiment Results and Analyses}
\begin{table}[t]
\begin{center}
\setlength{\tabcolsep}{1.6mm}
\fontsize{8.25}{11}\selectfont
\begin{tabular}{|c|c|c|c|c|}
\hline 
\textbf{Methods} & $\;$\textbf{AUC}$\;\;$ & $\;$\textbf{MRR}$\;\;$ & \fontsize{8}{11}\selectfont{\textbf{nDCG@5}} & \fontsize{8}{11}\selectfont{\textbf{nDCG@10}}\\
\hline
LibFM& 61.16 & 27.88 & 30.06 & 36.44\\
DSSM& 64.74 & 30.12 & 33.22 & 39.50\\
Wide\&Deep& 64.62 & 29.87 & 32.71 & 39.11\\
\hline
DAE-GRU& 65.98 & 31.48 & 34.93 & 41.12\\
DFM& 64.63 & 29.80 & 32.82 & 39.29\\
DKN& 66.20 & 31.25 & 34.23 & 40.92\\
LSTUR& 68.10 & 32.87 & 36.46 & 42.69\\
NAML& $\;\,${68.63}$^{\dagger}$ & 33.16 & 36.79 & 43.07\\
NPA& 67.34 & 32.59 & 35.98 & 42.28\\
NRMS& 68.61 & $\;\,${33.46}$^{\dagger}$ & $\;\,${37.02}$^{\dagger}$ & $\;\,${43.30}$^{\dagger}$\\
FIM& 68.44& 32.95& 36.58& 42.97\\
\hline
CNE-SUE& $\;\,$\textbf{69.55}$^{\star}$& $\;\,$\textbf{33.70}$^{\star}$& $\;\,$\textbf{37.54}$^{\star}$ & $\;\,$\textbf{43.79}$^{\star}$\\
\hline
CNE w/o CS& 69.39& 33.52& 37.30& 43.62\\
CNE w/o CA& 69.48& 33.61& 37.39& 43.68\\
SUE w/o GCN& 69.31& 33.48& 37.25& 43.53\\
SUE w/o HCA& 69.40& 33.52& 37.37& 43.65\\
\hline
\end{tabular} 
\end{center}
\captionsetup{font=10pt}
\caption{Performance comparison results~($^{\dagger}$ denotes the highest baseline scores, $^{\star}$ denotes that the performance improvements over all baseline methods are validated by Student’s unpaired t-test with $p$-value $<0.01$).}
\label{tables:main_experiments}
\end{table}
\subsection{Main Comparison Results}
Table~\ref{tables:main_experiments} shows the performance comparison results. Our model CNE-SUE achieves the highest performance consistently in all evaluation metrics. Detailed observations can be obtained as follows.

General recommendation methods yield much lower performance than most neural news recommendation methods. This is due to that deep neural models can learn refined representations adaptively, which are more effective than general feature engineering with fixed handcrafted features.

In all evaluation metrics, our model CNE-SUE outperforms all comparison methods by significant margins~($+0.92\%$ AUC, $+0.24\%$ MRR, $+0.52\%$ nDCG@$5$, and $+0.49\%$ nDCG@$10$ compared to the best baseline performance). This performance improvement derives from the collaborative news representations and structural user representations extracted by our model. Specifically, CNE can extract more accurate news semantics by leveraging the title-content semantic interaction, compared to title-encoding~(e.g., LSTUR, NRMS, and FIM) and separate encodings of title and content~(e.g., NAML). SUE modeling diverse user interests with hierarchical cluster structure is more powerful than the comparison methods formulating user history as a linear sequence of news, which employ recurrent neural networks~(e.g., LSTUR) or attention networks~(e.g., NPA, NAML, and NRMS).

From table~\ref{tables:main_experiments}, we can observe varying degrees of performance decreases on the ablation variants compared to our full model. It suggests the usefulness of different components in our model. CNE~w/o~CA performs the best among all variants. This is because the news representations learned by CNE are composed of self- and cross-attentive representations~(refer to Eq.~(\ref{eq8})), and the remaining self-attention can achieve suboptimal performance. Removing GCN layers leads to the most significant impact on performance, indicating the efficacy of structural modeling on user history.

\subsection{Effectiveness of Collaborative Encoding}\label{CNE_abltions}
\begin{table}[t]
\begin{center}
\setlength{\tabcolsep}{1.6mm}
\fontsize{8.25}{11}\selectfont
\begin{tabular}{|c|c|c|c|c|}
\hline 
\textbf{News Enc.} & $\;$\textbf{AUC}$\;\;$ & $\;$\textbf{MRR}$\;\;$ & \fontsize{8}{11}\selectfont{\textbf{nDCG@5}} & \fontsize{8}{11}\selectfont{\textbf{nDCG@10}}\\
\hline
CNN& 68.31 & 33.07 & 36.65 & 42.92 \\
KCNN& 65.27 & 30.72 & 33.67 & 40.28 \\
Per-CNN& 68.37 & 33.12 & 36.68 & 42.90\\
MHSA& 68.26 & 33.07 & 36.59 & 42.83 \\
NAML&  $\;\,${68.63}$^{\dagger}$ & $\;\,${33.16}$^{\dagger}$ & $\;\,${36.79}$^{\dagger}$ &  $\;\,${43.07}$^{\dagger}$\\
\hline
CNE& \textbf{69.21}& \textbf{33.32} & \textbf{37.16}& \textbf{43.44}\\
\hline
NAML-T& 68.49 & 33.21 & 36.89 & 43.11\\
NAML-C& 67.22 & 32.08 & 35.62 & 41.76\\
CNE-T& 68.25& 33.18& 36.80& 43.12\\
CNE-C& 67.96& 32.65 & 36.34& 42.44\\
\hline
\end{tabular} 
\end{center}
\captionsetup{font=10pt}
\caption{Ablation study of news encoders~(KCNN denotes knowledge-aware CNN in DKN, Per-CNN denotes CNN with personalized attention in NPA, MHSA denotes multi-head self-attention networks in NRMS, $*$-T(C) denotes title~(content) encoding only).}
\label{tables:news_ablations}
\end{table}
We conduct ablation experiments on news encoders. For fair comparison and excluding the influence of SUE, all ablation models apply the same attention user encoders. We also examine the title and content encodings. Table~\ref{tables:news_ablations} shows the ablation results\footnote{We did not include the ablation of FIM, because the FIM news encoder produces special hierarchical 3D-sized representations, which are incompatible with other ablation encoders.}.

From the ablation results, we can observe that CNE significantly outperforms other existing news encoding methods. NAML is also competitive, as it can incorporate informative representations from news texts and topic categories. Table~\ref{tables:news_ablations} also shows that title-encoding~($*$-T) is much more effective than content-encoding~($*$-C), though content texts are theoretically more informative. This confirms the ``\textit{semantic encoding dilemma}'' in news encoding and may explain why many existing works~(e.g., LSTUR, NPA, NRMS, and FIM) employ title-encoding only. Comparing NAML to NAML-T, there is no significant performance enhancement. In contrast, CNE achieves much higher scores than the individual title and content encodings~(CNE-T and CNE-C). It validates the necessity of encoding news title and content with word-level semantic interaction to enhance news representation learning.

\subsection{Effectiveness of Structural Encoding}\label{SUE_abltions}
\begin{table}[t]
\begin{center}
\setlength{\tabcolsep}{1.6mm}
\fontsize{8.25}{11}\selectfont
\begin{tabular}{|c|c|c|c|c|}
\hline 
\textbf{User Enc.} & $\;$\textbf{AUC}$\;\;$ & $\;$\textbf{MRR}$\;\;$ & \fontsize{8}{11}\selectfont{\textbf{nDCG@5}} & \fontsize{8}{11}\selectfont{\textbf{nDCG@10}}\\
\hline
LSTUR& 68.10 & 32.87 & 36.46 & 42.69\\
ATT& 68.31 & 33.07 & 36.65 & 42.92\\
Per-ATT& 66.97& 32.22& 35.48 & 41.82\\
Can-ATT& 68.48& 33.23& 36.73 & 43.04\\
MHSA& $\;\,${68.75}$^{\dagger}$ & $\;\,${33.34}$^{\dagger}$ & $\;\,${36.97}$^{\dagger}$ & $\;\,${43.26}$^{\dagger}$\\
\hline
SUE& \textbf{69.03} & \textbf{33.53} & \textbf{37.26} & \textbf{43.48}\\
\hline
\end{tabular} 
\end{center}
\captionsetup{font=10pt}
\caption{Ablation study of user encoders~(ATT denotes vanilla attention networks, Per-ATT denotes personalized attention in NPA, Can-ATT denotes candidate-aware attention in DKN, MHSA denotes multi-head self-attention networks in NRMS).}
\label{tables:user_ablations}
\end{table}
We conduct ablation experiments on user encoders. For fair comparison and excluding the influence of CNE, all ablation models apply the same CNN title encoders. Table~\ref{tables:user_ablations} shows the ablation results.

According to Table~\ref{tables:user_ablations}, MHSA performs much better than other baseline user encoders. This is because MHSA~\citep{transformer} can model the correlation of each pair of news in user history. It validates the necessity of modeling the correlation of historical news in user encoding. Moreover, Table~\ref{tables:user_ablations} shows that SUE significantly outperforms MHSA. This is because the manner of encoding historical news correlation with hierarchical clusters in SUE is more fine-grained than MHSA. Concretely, modeling intra-cluster news interaction is more effective to reflect aspects of user interests, while modeling inter-cluster user interests interaction is more effective to encode overall user representations. The ablation results indicate that structural modeling of the hierarchical \textit{user-interest-news} correlation can effectively enhance user encoding.

\begin{figure}[t]
\centering
\includegraphics[width=75.25mm]{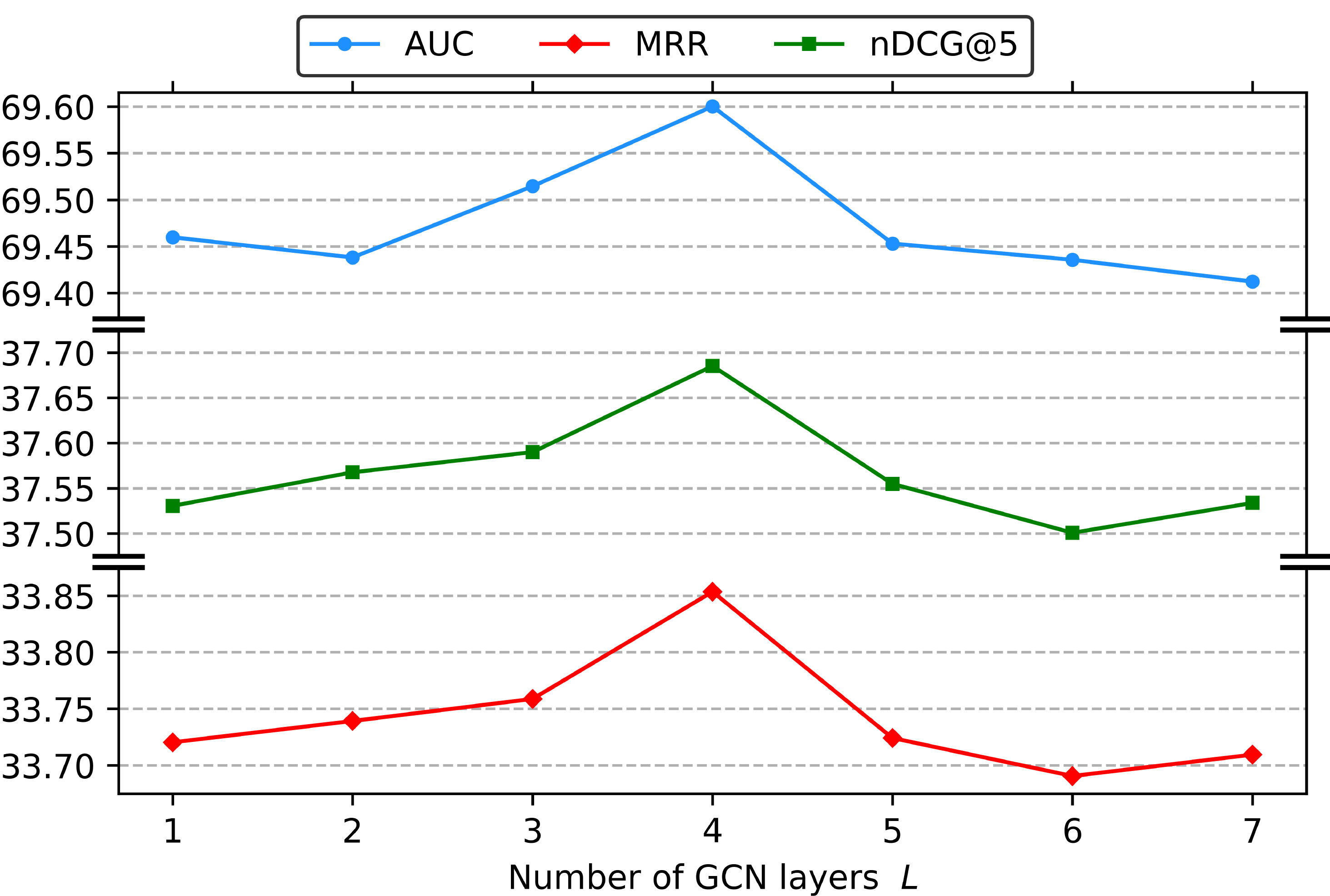}
\captionsetup{font=10pt}
\caption{
The performance of our model on validation set with respect to the number of GCN layers~(the trend of ndcg@10 is similar to ndcg@5 and hence omitted).
}
\label{fig:GCN_depth}
\end{figure}
\subsection{Parameter Analysis}\label{GCN_layer_num}
We investigate the influence of the number of GCN layers $L$ in our model. Figure~\ref{fig:GCN_depth} shows the results. The model performance on validation set increases and reaches a peak, as $L$ increases from $1$ to $4$. This is because equipped with deeper GCN, the model can capture more fine-grained information of user browsing behaviors by modeling higher-order interaction of browsed news. Nevertheless, as $L$ continues to increase, the model performance begins to decline. This may be because deep GCN always suffers from the \textit{over-smoothing} issue~\citep{over-smoothing}. As GCN becomes too deep, the user history representations $r^h$ tend to be indistinguishable and impair the ultimate user representations. Herein, $L=4$ is optimal for our model.

\begin{figure}[t]
\centering

\subfigcapskip=-1.1mm
\subfigure[Attention weights of our model over the title words.]{
\includegraphics[width=74mm, fbox]{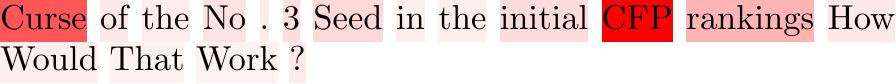}\label{subfig:title_visualization}
}
\vskip -0.3mm
\subfigure[Attention weights of our model over the content words.]{
\includegraphics[width=74mm, fbox]{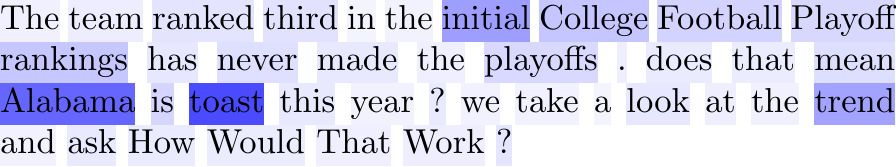}\label{subfig:content_visualization}
}
\captionsetup{font=10pt}
\caption{
Attention weight visualization on the news $N6$~(darker colors denote higher attention weights).
}
\label{fig:visualization}
\end{figure}
\subsection{Case Study}
We then probe into how our model processes news texts. According to Eq.~(\ref{eq6}), (\ref{eq7}), and (\ref{eq8}), we define the word attention weights of title~(content) as $\alpha^{t(c)}=(\alpha^{t(c)}_{self}+\alpha^{t(c)}_{cross})/2$, where $\alpha^{t(c)}\in{[0,1]}$. As shown in Figure~\ref{fig:visualization}, we visualize our model's output title~(content) attention weights $\alpha^{t(c)}$ over the title~(content) words of the news $N6$ in Figure~\ref{fig:news_user}.

From Figure~\ref{subfig:title_visualization}, we observe that our model mainly attends to the words ``\textit{curse}'' and ``\textit{CFP}'', which contain the core information of the news $N6$. It validates that our model can distill the most informative words from the news title. As the content visualization shown in Figure~\ref{subfig:content_visualization}, our model mostly attends to the words ``\textit{Alabama}'' and ``\textit{toast}'', which interpret the specific semantics of the word ``\textit{curse}'' in the context of the news $N6$. Besides, our model also attends to the important contextual words, such as ``\textit{initial}'', ``\textit{rankings}'', and ``\textit{trend}''. These title and content attention weights indicate that our model can accurately encode the news $N6$.

\begin{figure}[t]
\centering
\includegraphics[width=75mm]{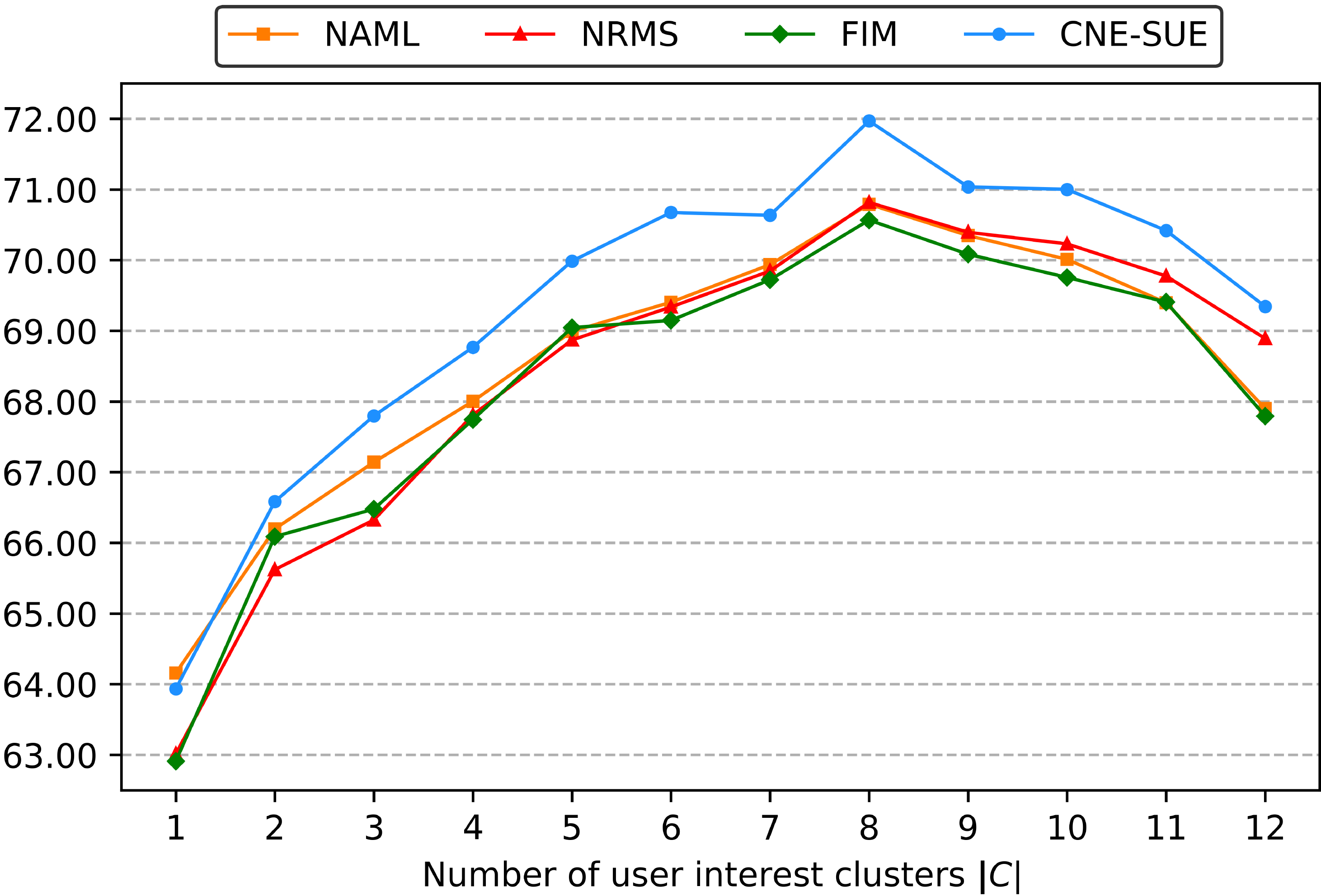}
\captionsetup{font=10pt}
\caption{
The AUC scores of different models with respect to the number of user interest clusters.
}
\label{fig:cluster_num}
\end{figure}
\subsection{Analysis on User Interest Modeling}
We analyze how our model performs with different numbers of user interest clusters $|C|$~(refer to Section~\ref{ssec:SUE}). The results are shown in Figure~\ref{fig:cluster_num}. When $|C|=1$, our model slightly underperforms NAML. This is because our model is overfitted to represent single user interest with cluster graphs. All models' performance increases with the growth of $|C|$. This is because the models can learn more precise user representations as more news information is incorporated. In cases of $|C|>1$, our model consistently outperforms all baselines. It validates the usefulness of encoding user history with hierarchical cluster-structure in cases of modeling diverse user interests. Moreover, the performance of all models decreases when $|C|$ becomes too large~(i.e., $|C|>8$). This reveals the challenge of predicting a user's news-clicking behavior when her browsing history covers too many kinds of news topics.

\section{Conclusion}
In this work, we present a neural news recommendation model with collaborative news encoding and structural user encoding. CNE leverages the title-content semantic interaction to enhance news encoding. SUE exploits the correlation of browsed news and represents user interests with hierarchical cluster graphs to enhance user encoding. Experiment results show that our model achieves significant performance enhancement compared to the existing state-of-the-art methods. We also further analyze our model and validate its effectiveness.

\section*{Acknowledgements}
We appreciate some insightful comments from the anonymous reviewers. The research described in this paper is partially supported by Hong Kong RGC-GRF \#14204118 and Hong Kong RSFS \#3133237.
\bibliography{anthology,custom}
\bibliographystyle{acl_natbib}
\end{document}